\documentclass{article}

\usepackage{arxiv}

\usepackage[utf8]{inputenc} % allow utf-8 input
\usepackage[T1]{fontenc}    % use 8-bit T1 fonts
\usepackage{hyperref}       % hyperlinks
\usepackage{url}            % simple URL typesetting
\usepackage{booktabs}       % professional-quality tables
\usepackage{amsfonts}       % blackboard math symbols
\usepackage{nicefrac}       % compact symbols for 1/2, etc.
\usepackage{microtype}      % microtypography
\usepackage{lipsum}		% Can be removed after putting your text content
\usepackage{graphicx}
\usepackage{natbib}
\usepackage{doi}

\usepackage{xcolor}
\usepackage{multirow}
\usepackage{float}
\usepackage{caption}
\usepackage{subcaption}

\title{Exploiting pretrained biochemical language models for
targeted drug design}

\author{ \href{https://orcid.org/0000-0002-8684-2457}{\includegraphics[scale=0.06]{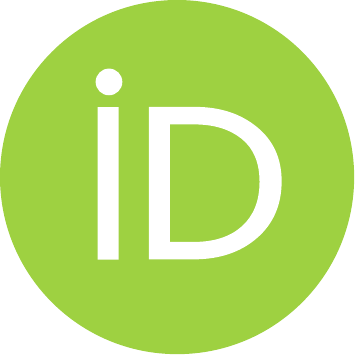}\hspace{1mm}Gökçe Uludoğan}
\\
	Department of Computer Engineering\\
	Bogazici University\\
	Istanbul 34342, Turkey \\
	\texttt{gokce.uludogan@boun.edu.tr} \\
	\And
	\href{https://orcid.org/0000-0002-3206-8427}{\includegraphics[scale=0.06]{orcid.pdf}\hspace{1mm}Elif Ozkirimli} \\
	Data and Analytics Chapter\\
	Pharma International Informatics\\
	F. Hoffmann-La Roche AG 4303, Switzerland \\
	\texttt{elif.ozkirimli@roche.com} \\
	\AND
	 \href{https://orcid.org/0000-0003-3668-3467}{\includegraphics[scale=0.06]{orcid.pdf}\hspace{1mm}Kutlu O. Ulgen} \\
	 Department of Chemical Engineering \\
    Bogazici University\\
	Istanbul 34342, Turkey \\
	\texttt{ulgenk@boun.edu.tr} \\
	 \And
	 \href{https://orcid.org/0000-0002-6916-122X}{\includegraphics[scale=0.06]{orcid.pdf}\hspace{1mm}Nilgün Karalı} \\
    Department of Pharmaceutical Chemistry\\
    Istanbul University \\
     Istanbul 34116, Turkey \\
	 \texttt{karalin@istanbul.edu.tr} \\
	 \And
     \href{https://orcid.org/0000-0001-8376-1056}{\includegraphics[scale=0.06]{orcid.pdf}\hspace{1mm}Arzucan Özgür} \\
	Department of Computer Engineering\\
	Bogazici University\\
	Istanbul 34342, Turkey \\
	 \texttt{arzucan.ozgur@boun.edu.tr} \\
}

\date{}

\renewcommand{\shorttitle}{Exploiting Pretrained Biochemical Language Models for Targeted Drug Design}

\hypersetup{
pdftitle={Exploiting pretrained biochemical language models for targeted drug design},
pdfsubject={q-bio.BM, q-bio.QM, cs.CL, cs.LG, stat.ML},
pdfauthor={Gökçe Uludoğan, Elif Ozkirimli, Kutlu O. Ulgen, Nilgün Karalı, Arzucan Özgür},
pdfkeywords={deep generative model, pretrained biochemical language models, targeted drug design},
}

\begin{document}
\maketitle
\def\thefootnote{*}\footnotetext{This article has been accepted for publication in \textit{Bioinformatics} Published by Oxford University Press.}\def\thefootnote{\arabic{footnote}}

\begin{abstract}
\textbf{Motivation:} The development of novel compounds targeting proteins of interest is one of the most important tasks in the pharmaceutical industry. Deep generative models have been applied to targeted molecular design and have shown promising results. Recently, target-specific molecule generation has been viewed as a translation between the protein language and the chemical language. However, such a model is limited by the availability of interacting protein-ligand pairs. On the other hand, large amounts of unlabeled protein sequences and chemical compounds are available and have been used to train language models that learn useful representations. In this study, we propose exploiting pretrained biochemical language models to initialize (i.e. warm start) targeted molecule generation models. We investigate two warm start strategies: (i) a one-stage strategy where the initialized model is trained on targeted molecule generation (ii) a two-stage strategy containing a pre-finetuning on molecular generation followed by target specific training. We also compare two decoding strategies to generate compounds: beam search and sampling.

\textbf{Results:} The results show that the warm-started models perform better than a baseline model trained from scratch. The two proposed warm-start strategies achieve similar results to each other with respect to widely used metrics from benchmarks. However, docking evaluation of the generated compounds for a number of novel proteins suggests that the one-stage strategy generalizes better than the two-stage strategy. Additionally, we observe that beam search outperforms sampling in both docking evaluation and benchmark metrics for assessing compound quality.

\textbf{Availability and implementation:} The source code is available at \url{https://github.com/boun-tabi/biochemical-lms-for-drug-design} and the materials (i.e., data, models, and outputs) are archived in Zenodo at \url{https://doi.org/10.5281/zenodo.6832145}.

%\textbf{Contact:} arzucan.ozgur@boun.edu.tr or elif.ozkirimli@roche.com
\end{abstract}

% keywords can be removed

\section{Introduction}

The discovery of molecules with desired properties is a crucial task in drug discovery. To find molecules biased towards a biological target, candidate compounds must be identified from the vast space of potential drug-like molecules, estimated to be in the order of $10^{60}$ \citep{polishchuk2013estimation}. Although advances in high-throughput screening allow screening of a large number of molecules against a biological target, these experiments are expensive, and the space is still too large for screening. Another challenge arises from the promiscuity of small drug molecules, i.e., their ability to interact with many targets. Drugs designed for one target have been shown to be active against an average of 11 other targets \citep{peon2017predicting}. The promiscuity of molecules could lead to unexpected toxicity, one of the primary causes of attrition in drug discovery \citep{kramer2007application}. These issues and the complexity of human biology present a challenge and an opportunity for the development of scalable and efficient drug discovery tools.

Over the last decades, computational methods have been utilized
to accelerate drug discovery and minimize costs. More recently, deep generative models have been used for de novo drug design \citep{aumentado2018latent,Skalic2019,Moret2020,Born2021,Chenthamarakshan2020}. These models are data-driven and scalable, and also allow efficient navigation in chemical space. Such generative models can be broadly classified into two groups based on the usage of the target information. The first group of models utilizes
known compounds to guide generation. In this line of work, the models are first trained with a large number of molecules to learn the chemical space. Later, they are fine-tuned with several approaches such as transfer learning, reinforcement learning, and Bayesian optimization for biased molecule generation. However, these approaches either require molecules with activity against the target of interest, which are not available for novel proteins, or a predictive model that biases generation, which is shown to be
problematic \citep{renz2020failure}. Another line of work consists of models leveraging the structure of the target. However, the major limitation of these models is the scarcity of structural information for targets. Contrary to these studies, \citet{Grechishnikova2021} proposed using the sequences of proteins
for target-specific molecule generation. The study viewed the targeted molecule generation problem as a translation task between protein and molecular languages and adopted the transformer architecture. The major limitation of this study is the training data set, which consists of 200K protein-ligand interactions belonging to around 1000 proteins. Compared to the vast amounts of unlabeled sequences available, the data set is small, limiting the model’s generalizability.

Lately, self-supervised pretraining has become the dominant paradigm in natural language processing (NLP). Self-supervised methods exploit large amounts of unlabeled data for pretraining. The pretrained models are then fine-tuned on another task to save computational time and benefit from representations learned from large amounts of data. This paradigm has obtained state-of-the-art results in many NLP tasks \citep{Devlin2019}. The rapidly growing data of biochemical entities and the analogies between human languages and biochemical languages have enabled the application of these methods in bio/cheminformatics \citep{ozccelik2021chemboost, Chithrananda, Filipavicius2020, rives2021biological, elnaggar2020prottrans}. By viewing protein sequences as sentences and one or several amino acid residues as words, these techniques have been adopted in protein language modelling and shown to learn useful representations and improve the state-of-the-art in a number of protein engineering tasks \citep{rives2021biological}. Similarly, by treating the Simplified Molecular Input Line Entry System (SMILES) \citep{weininger1988smiles} string of each molecule as a sentence and each symbol or subsequent symbols as a word, chemical language models have been trained and shown to obtain
promising results on downstream tasks \citep{Chithrananda}.

We view target-specific molecule generation as a translation task
from protein language to chemical language. To overcome the challenge of limited data, we propose exploiting biochemical language models pretrained on large scale sequences to initialize (i.e., warm start) target-specific molecule generation models. We hypothesize that warm starting will allow the model to benefit from the representations learned by the pretrained models on diverse sequences, thereby enhancing generalizability and boosting performance.

We employ a Transformer based sequence-to-sequence model \citep{rothe2020leveraging} and initialize the encoder and the decoder components with pretrained models on large-scale data. We investigate two warm start strategies: one-stage strategy where the initialized model is fine-tuned on protein-ligand pairs filtered from BindingDB \citep{Gilson2016}, and two-stage strategy including an initial fine-tuning with compounds from MOSES \citep{Polykovskiy2020} followed by training with interacting protein-compounds pairs. To compare with the warm-started models,
we adopt the T5 Transformer model \citep{Raffel2019}. The results
show that the proposed approach is effective, as the warm-started models achieved better performance compared to the model trained from scratch. The introduced warm-start strategies perform similarly to each other with respect to the metrics of the MOSES benchmark. However, docking evaluation of the designed compounds for a set of novel proteins suggests that the one-stage strategy generalizes better than the two-stage strategy. We also compare two decoding strategies, beam search and sampling, for generating molecules, and show that beam search outperforms sampling based on both the docking evaluation and the benchmark metrics for assessing the quality of generated chemical molecules.

\section{Materials and Methods}
\label{sec:methods}

\subsection{Data}
To train and assess targeted generative models, we used protein-ligand interactions filtered from BindingDB \citep{Gilson2016} which contains measured binding affinities between proteins and small molecules. First, we filtered out the interactions missing either protein or SMILES sequence. In 95\% of the reported interactions, the proteins were assayed a single
chain, so we dropped multichain proteins for simplicity. We used UniProt identifiers of proteins to obtain families from Pfam \citep{mistry2021pfam}; thus, we also excluded the proteins without UniProt identifier or Pfam family. After omitting the interactions not having any affinity measurement, the remaining interactions were labelled as active or inactive based on affinity scores. BindingDB includes several affinity metrics ($K_i$, $K_d$, $IC_{50}$, $EC_{50}$) which are not directly comparable. $K_i$ and $K_d$ values are first converted to $IC_{50}$ by a factor 2 by following \citep{jansson2020does} to set an affinity threshold. Then, we labelled the interactions with affinity scores below 100 nm as active, and those with affinity scores above 10000 nm as inactive \citep{gao2018interpretable}. For the protein-ligand pairs with
multiple reported affinity scores and assay, we calculated the geometric mean of these values and compared this score with the thresholds to label these interactions \citep{jansson2020does}. The statistics of the resulting data set are reported in in Table \ref{data:bdb}.

\begin{table}[H]
    \centering
  \caption{\ Statistics of the data set extracted from BindingDB}
  \label{data:bdb}
    \begin{tabular}{llll}
    \toprule
    \textbf{Label}    & \textbf{\# Interactions} & \textbf{\# Unique Proteins} & \textbf{\# Unique Ligands} \\ 
    \midrule
    \textbf{Active}   & 428067                   & 3099                        & 331942                     \\
    \textbf{Inactive} & 64696                    & 3817                        & 123763        \\ \bottomrule
    \end{tabular}
\end{table}

We relied on sequence similarities and Pfam families of proteins while splitting active interactions. The similarity of protein sequences were computed using Needleman-Wunsch global alignment with BioPython \citep{10.1093/bioinformatics/btp163} wrapper of EMBOSS \citep{rice2000emboss}. We aimed to create diverse validation and test sets. To this end, we sampled 10\% of proteins in each Pfam family and computed similarities between sampled
proteins and the remaining proteins. Then, the selected proteins were binned based on maximum similarities to the remaining proteins. From these bins, 200 proteins were chosen in total with weighted random sampling based on the inverse frequency of the bins. The interactions of these proteins constitute the validation set. The steps described above were repeated with the remaining proteins to form the test set. Distributions of sequence similarities between and within these splits are shown in
Figure \ref{fgr:seq_sim}. The summary of the splits is presented in Table \ref{data:splits}.

\begin{table}[H]
    \centering
  \caption{\ Summary of splits}
  \label{data:splits}
    \begin{tabular}{llll}
    \toprule
    \textbf{Split}    & \textbf{\# Interactions} & \textbf{\# Unique Proteins} & \textbf{\# Unique Ligands} \\ 
    \midrule
    \textbf{Training}   & 310300 & 2337 & 257348                     \\
    \textbf{Validation} & 25335 & 200 & 24121  \\
    \textbf{Test}  & 21350 & 200 & 20675 \\
    \bottomrule
\end{tabular}
\end{table}

Aside from BindingDB data, we also used MOSES dataset \citep{Polykovskiy2020} filtered from the ZINC database \citep{sterling2015zinc} since the two stage warm start strategy requires a collection of compounds. This dataset contains 1,936,962 drug-like molecules split into training (1.6M), test (176K) and scaffold tests (176K). 

\subsection{Representation}
The view of targeted drug design as a translation task relies on the analogies between biochemical languages (i.e., protein and chemical languages) and human languages. Proteins and chemical compounds can be represented as sequences like sentences in human languages. Furthermore, proteins are composed of functional units shared between different proteins. Likewise, chemical compounds consist of subfragments that have been shown to follow power-law similarly to natural languages \citep{wozniak_linguistic_2018}. Numerious studies adopted methods from natural language processing
(NLP) to process biochemical sequences and identify meaningful units of these languages \citep{ozccelik2021chemboost,Chithrananda,Filipavicius2020,rives2021biological,elnaggar2020prottrans}. Recent studies have applied subword segmentation for splitting sequences into tokens and have shown that this approach performs better compared to character-level tokens \citep{li2021smiles,asgari_probabilistic_2019}. Therefore, we consider protein and chemical units identified by subword segmentation algorithms as tokens rather than individual characters, unlike the previous work \citep{Grechishnikova2021}. This reduces the length of the sequences and allows better capturing of long-range dependencies, which is crucial for this problem. We used the vocabularies constructed with the Byte Pair Encoding (BPE) algorithm (Sennrich et al., 2015) by the pretrained models, which we employed for warm starting. The protein vocabulary contains 10K protein words, while the chemical vocabulary comprises 8K chemical words.

\subsection{Models}
We applied a Transformer-based sequence-to-sequence model on target-specific molecule generation with warm starting from the pretrained models. The model is based on encoder-decoder architecture, the prevalent approach for sequence-to-sequence tasks in NLP. In this paradigm, the encoder takes an input sequence and encodes it to a sequence of hidden states while the decoder produces an output sequence autoregressively given the encoder’s hidden states. We used pretrained Transformer variants for initializing our model to benefit from the representations learned on diverse sequences.

The original Transformer is introduced by \citet{Vaswani2017} and shown to be effective in various NLP tasks and adapted in different domains \citep{Mahmood2020,Fabian,Chithrananda,Filipavicius2020,Devlin2019}. The key component of this model is self-attention which allows the model to relate different parts of the sequence while computing the representation of the sequence. Relying on attention blocks makes the model computationally efficient and helps capture long-range dependencies. The model is composed of encoder and decoder stacks containing self-attention and feed-forward layers. The self-attention mechanism in the encoder and the decoder is similar; however, the decoder uses a causal attention mask to prevent attending to the next words for autoregressive generation. In addition, the decoder has cross attention layers that attend to the encoder’s hidden states.

In this study, we leveraged Transformer variants pretrained on large datasets to initialize models. For the encoder part, we utilized the checkpoints of Protein RoBERTa \citep{Filipavicius2020}. Protein RoBERTa model is pretrained with the masked language modeling on 5M binding/non-binding protein sequence-pairs collected from the STRING database \citep{Szklarczyk2019}. Contrary to other pretrained protein
models, this model represents proteins with subwords identified by the BPE algorithm, and thus, supports long sequences without increasing memory requirements. These properties of the model align with our problem formulation. Protein RoBERTa uses the encoder part of the Transformers, and all encoder parameters required by our model can be transferred from this model. To initialize the decoder of our model, we adopted the ChemBERTa model \citep{Chithrananda} pretrained on 10M PubChem compounds \citep{Kim2019} with BPE tokenization. The model does not achieve state-of-the-art performance; however, it has been shown to learn better representations as is trained with additional data. Similar to Protein RoBERTa, ChemBERTa also uses RoBERTa model \citep{liu2019roberta} based on Transformer encoder and is pretrained with masked language modeling. Since ChemBERTa consists of only encoder blocks, ChemBERTa weights can be used to initialize self-attention and feed-forward layers of the decoder only. The cross attention (i.e., encoder-decoder attention) and language model head (i.e., the layers mapping representations to output word probabilities) are initialized randomly.

We proposed two warm start strategies for training the encoder-decoder model. The first strategy is one stage finetuning where the model weights are initialized with Protein RoBERTa and ChemBERTa checkpoints, and then, the warm-started model is trained on protein-ligand interactions filtered from BindingDB. We refer to the model trained with this strategy as EncDecBase. Two-stage finetuning is another warm start strategy, including an initial finetuning of ChemBERTa on MOSES dataset to obtain a generic molecule generator, namely ChemBERTaLM. In this strategy, ChemBERTaLM is used to initialize the decoder instead of the original ChemBERTa. Then the initialized model is trained with interacting protein-ligand pairs similar to the one-stage strategy. We call this model EncDecLM. 

\begin{figure}[h]
    \centering
    \includegraphics[width=\textwidth]{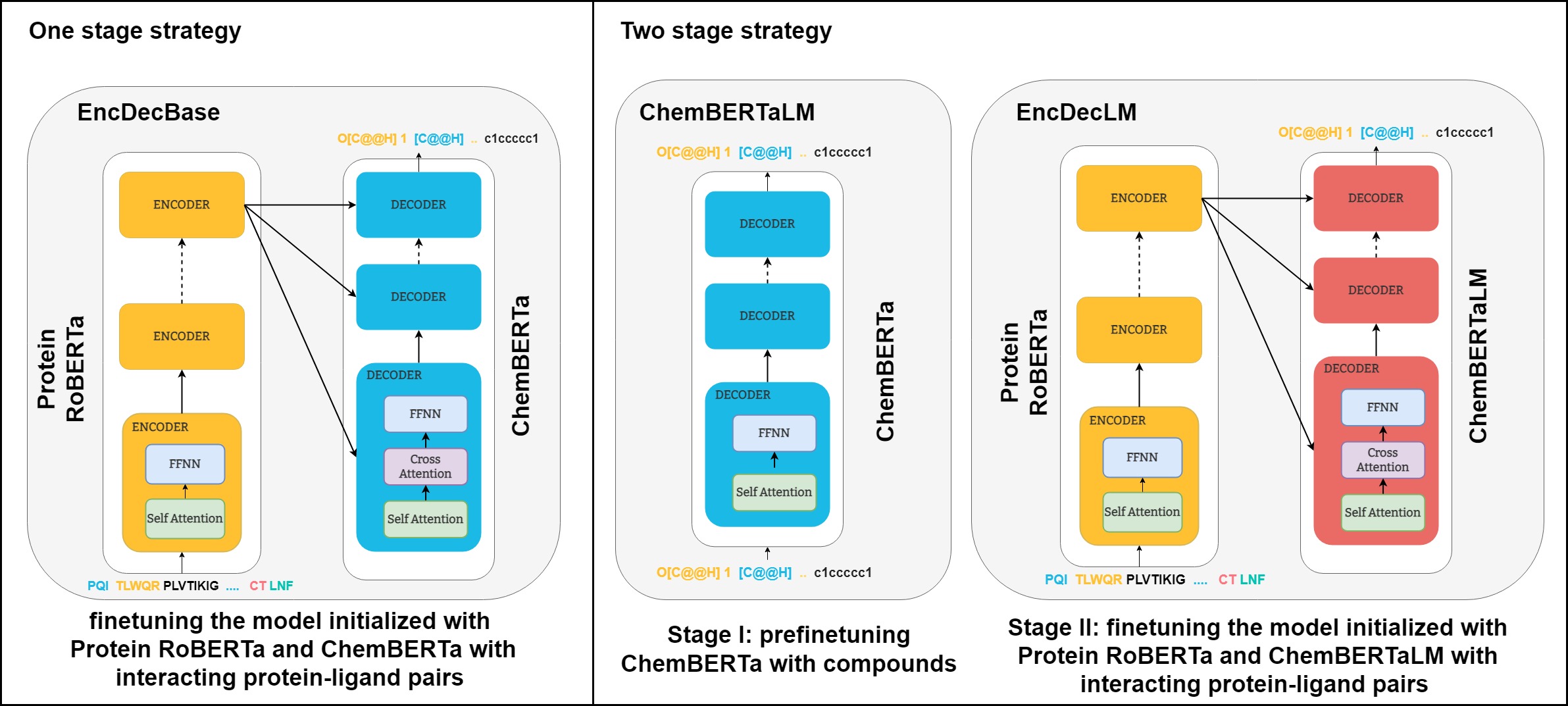}
    \caption{Warm start strategies}
    \label{fig:strategies}
\end{figure}

To compare with the warm-started models, the T5 Transformer model is used \citep{Raffel2019}. This model achieved state-of-the-art results in generation tasks and differs slightly from the original Transformer, which is also used in previous work for protein-specific molecule generation \citep{Grechishnikova2021}. Recent studies showed that training bigger models with early stopping is the best compute-efficient training strategy. For this reason, we build a slightly larger model compared to those in the previous work. Our T5 model uses four layers with a hidden size of 256, a feed-forward layer size of 512, and 6 attention heads.

We implemented all models using HuggingFace’s transformers library \citep{Wolf2020}. ChemBERTaLM was finetuned for 10 epochs with default settings. The warm started models were finetuned using Adam optimizer with a linear learning rate schedule with 2000 warm-up steps followed by a linear decay. The batch size was set to 8, and the gradients were accumulated every 8 steps to obtain an effective batch size of 64. Encoder decoder models (i.e., EncDecBase and EncDecLM) were trained
for 5 epochs. On the other hand, the T5 model was trained from scratch and needed more epochs to converge. This model was trained for 10 epochs. Once the models were trained, the checkpoints with the best validation performance were used to generate molecules.

\subsection{Evaluation}
\subsubsection{Benchmarking metrics}
We used a set of metrics from MOSES benchmark to assess the quality
and diversity of the generated chemical compounds.

\begin{itemize}
    \item \textbf{Validity:} The percentage of valid compounds. A compound is considered valid if it can be parsed by RDKit \citep{Landrum2016RDKit2016_09_4}. 
    \item \textbf{Uniqueness:} The percentage of unique molecules.
    \item \textbf{Novelty:} The percentage of novel compounds (i.e., not present in the training set).
    \item \textbf{Fréchet ChemNet distance (FCD) \citep{Preuer2018}:} A distance metric computing chemical and biological similarity between two sets of compounds.
    \item \textbf{Scaffold Similarity:} Cosine similarity of Bemis-Murcko scaffold \citep{Bemis1996} frequencies between two groups of compounds.
    \item \textbf{Fragment Similarity:} Cosine similarity between frequencies of BRICS fragments \citep{degen2008art} of two compound sets.
    \item \textbf{Nearest Neighbor Tanimoto Similarity (SNN):} The average Tanimoto similarity between a set of compounds and their nearest neighbours in another set of compounds.
    \item \textbf{Internal Diversity:} the average Tanimoto distance within a set of compounds where Tanitomo distance is equal to 1 minus Tanimoto similarity. This metric can be computed with powers of Tanimoto similarity. 
    % $IntDiv_{2}$ refers to the average squared Tanimoto distance between compounds. 
    \item \textbf{Filters:} The percentage of compounds passing a set of filters.
\end{itemize}

\subsubsection{Docking}
To assess the generated chemical compounds in terms of binding affinity, we performed docking for a number of novel proteins. To measure the discriminative ability of the docking tool between groups of compounds, we used Receiver Operating Characteristic (ROC) and the corresponding area under the curve (AUC).

\section{Results}
\label{sec:resuls}
\subsection*{ChemBERTaLM outperforms baseline models}

ChemBERTaLM model was trained for warm-starting the targeted generative model. To assess ChemBERTaLM, we generated 30K molecules by sampling and benchmarked the model using MOSES. Table \ref{tbl:moses} reports the performance of ChemBERTaLM model along with baseline models in the benchmark. 

\begin{table}[H]
\small
  \caption{Performance metrics for ChemBERTaLM and baseline models: fraction of valid molecules, fraction of molecules passing filters, fraction of novel molecules, internal diversity, Fréchet ChemNet Distance (FCD), Similarity to the nearest neighbor (SNN), Fragment similarity (Frag) and Scaffold similarity (Scaff). Test and TestSF denote MOSES random test set and scaffold split set respectively.}
  \label{tbl:moses}
  \resizebox{\textwidth}{!}{
\begin{tabular}{lllll|llll|llll}
\toprule
\multirow{2}{*}{\textbf{Model}} & \multirow{2}{*}{\textbf{Valid}} & \multirow{2}{*}{\textbf{Filters}} & \multirow{2}{*}{\textbf{Novelty}} & \multirow{2}{*}{\textbf{IntDiv}} & \multicolumn{4}{c|}{\textbf{Test}}                                            & \multicolumn{4}{c}{\textbf{TestSF}}                                          \\
                                &                                 &                                       &                                      &                                       & \textbf{FCD (↓)} & \textbf{SNN (↑)} & \textbf{Frag  (↑)} & \textbf{Scaf (↑)} & \textbf{FCD (↓)} & \textbf{SNN (↑)} & \textbf{Frag  (↑)} & \textbf{Scaf (↑)} \\ \midrule
Train                           & 1                               & 1                                                                     & 1       & 0.857                          & 0.008            & 0.642            & 1                  & 0.991              & 0.476            & 0.586            & 0.999              & 1                  \\ \midrule
AAE                             & 0.937                           & 0.996                        & 0.793         & 0.856                                                                 & 0.556            & 0.608            & 0.991              & 0.902              & 1.057            & 0.568            & 0.990              & 0.079              \\
CharRNN                         & 0.975                           & 0.994         & 0.842                           & 0.856                                                             & 0.073            & 0.601            & 1                  & 0.924              & 0.520            & 0.565            & 0.998              & 0.110              \\
VAE                             & 0.977                           & 0.997                         & 0.695          & 0.856                                                               & 0.099            & 0.626            & 0.999              & 0.939              & 0.567            & 0.578            & 0.998              & 0.059              \\
LatentGAN                       & 0.897                           & 0.973           & 0.949                           & 0.857                                                            & 0.296            & 0.538            & 0.999              & 0.886              & 0.824            & 0.514            & 0.998              & 0.100              \\
JTN-VAE                         & 1                               & 0.978              & 0.914                       & 0.851                                                             & 0.422            & 0.556            & 0.996              & 0.892              & 0.996            & 0.527            & 0.995              & 0.100              \\
ChemBERTaLM                       & 0.991                           & 0.997            & 0.844                       & 0.855                                                               & 0.090            & 0.609            & 1                  & 0.917              & 0.515            & 0.572            & 0.998              & 0.101         \\ \bottomrule     
\end{tabular}}
\end{table}

ChemBERTaLM outperformed the baseline models in terms of validity and performs on par in terms of the fraction of compounds passing filters and internal diversity. Although the model showed a lower novelty score than LatentGAN and JTN-VAE, it obtained the best FCD score on the scaffold split test set, which aims to assess generalizability of the model. ChemBERTaLM was also comparable to the baselines with respect to other similarity metrics (i.e. SNN, Frag, Scaf) on the test sets. These results indicate ChemBERTaLM generalizes well. We should also note that ChemBERTaLM model was fine-tuned for only 10 epochs while the baseline models were trained for 80-120 epochs.

\subsection*{Warm started models perform the best or on par across
metrics}

To investigate the utility of the warm start strategies, we trained the targeted generative models (i.e., EncDecBase, EncDecLM and T5). Then, we generated 20 molecules for each protein in the test set with two decoding methods, beam search and sampling. To assess the model performances, we first computed a set of metrics assessing the feasibility of the generated molecules and their similarity to the molecules in the test set. The results are shown in Table \ref{tbl:targeted}. The warm-started models outperformed T5 model trained from scratch. EncDecLM model performed the best or on par with EncDecBase across metrics, suggesting the utility of initial finetuning on molecules. However, the gain obtained by the pre-finetuning was limited, especially in terms of the similarity metrics (i.e., FCD, Scaf, SNN). EncDecBase model outperformed EncDecLM model with respect to the scaffold similarity (Scaf) and the nearest neighbour Tanimoto similarity (SNN).

\begin{table}[H]
\small
  \caption{\ Performance metrics for target specific generative models across decoding strategies}
  \label{tbl:targeted}
  \begin{tabular*}{\textwidth}{@{\extracolsep{\fill}}clcccccc}
    \toprule
    \textbf{Decoding}  & \textbf{Model} & \textbf{Valid} & \textbf{Unique} & \textbf{Novel} & \textbf{FCD (↓)} & \textbf{Scaf (↑)} & \textbf{SNN (↑)} \\
    \midrule
   \multirow{3}{*}{\textbf{Beam Search}}    & EncDecLM   & 0.984 & 0.795 & 0.965 & 9.454  & 0.090 & 0.560 \\
                                                  & EncDecBase & 0.961 & 0.780 & 0.978 & 11.652 & 0.100 & 0.572 \\
                                                  & T5         & 0.862 & 0.909 & 0.999 & 19.520 & 0.043 & 0.506 \\ \midrule
\multirow{3}{*}{\textbf{Sampling}}  & EncDecLM   & 0.908 & 0.967 & 0.992 & 4.519  & 0.088 & 0.413 \\
                                                  & EncDecBase & 0.840 & 0.986 & 0.996 & 4.366  & 0.085 & 0.389 \\
                                                 &  T5         & 0.664 & 1.000 & 1.000 & 4.688  & 0.032 & 0.341 \\
\bottomrule
    
  \end{tabular*}
\end{table}

The lowest FCD scores were obtained with the molecules generated
by sampling. This means that the sampling method produces more diverse molecules compared to the beam search.

To further investigate the performance of the models, we computed the evaluation metrics at the protein level. We computed FCD score between the generated compounds and the interacting compounds (i.e., those labeled as active) for each protein in the test set. For reference, we selected 20 interacting compounds for each protein in the training set and computed FCD score between the selected compounds and the rest of the active compounds of the corresponding protein. The distribution of FCD scores across the decoding strategies are shown in Figure \ref{fgr:fcd2}. The results show that the warm-started models achieved lower FCDs than the T5 model at the target level. We also followed the same procedure to compute SNN scores for the test proteins. SNN distributions also confirm that the warm started models perform better than the T5 model (see Figure \ref{fgr:snn}).

To compare the models at the protein level quantitatively, we computed Jensen-Shannon divergence (JSD) between FCD scores of the compounds generated by the models and those of the compounds interacting with the train proteins. The results are presented in Table \ref{tbl:jsd}. We observed that the compounds decoded with the beam search obtained a lower JSD score for the warm started models, while the compounds decoded with the sampling strategy resulted in lower JSD for the T5 models.

\begin{table}[H]
\caption{Jensen Shannon distance between FCD scores of the compounds generated for the test proteins and the scores of the compounds interacting with the train proteins.}
\label{tbl:jsd}
\begin{center}
\begin{tabular}{ccc}
\hline
\textbf{Model} & \textbf{Beam} & \textbf{Sampling} \\ \hline

EncDecBase   & \textbf{0.546} & 0.592            \\
                     %& 
                     EncDecLM       & 0.557         & \textbf{0.573}    \\
                     %&
                     T5             & 0.694         & 0.683             \\ \hline
\end{tabular}
\end{center}
\end{table}

\subsection*{Warm started models can generate target specific molecules}

Next, we performed docking to assess if generated chemical compounds are active against proteins of interest. The target proteins were selected randomly from the test proteins with at least 50 interacting compounds and a 3D structure in complex with a ligand. Given that docking is a computationally costly process, we limited the number of proteins to 12 for docking evaluation. For each protein, we selected one structure from PDB \citep{Burley2019}. The PDB codes for the selected proteins are as follows: 1ERE, 2WO6, 4B6L, 4I23, 5K0K, 5TUY, 5V1B, 5XY1, 6LVL, 6WJ5, 6Z1Q, 2PJL.

To test whether the generated compounds have binding affinity to the target protein of interest, we compiled three groups of compounds for each protein: i) the compounds having activity towards the target protein, ii) the compounds randomly selected from BindingDB, iii) the generated molecules for the target protein. Due to the technical limitations, a maximum of 100 compounds are included in each group. 

The ligands in each group were docked into corresponding protein structures with GNINA \citep{McNutt2021} which is a molecular docking tool based on deep learning. To this end, we extracted bound ligands from PDB structures using PyMol \citep{PyMOL}. The extracted ligands were used to define the binding site of the target for docking. We used RDKit \citep{Landrum2016RDKit2016_09_4} to generate conformers for the chemical compounds. Docking was performed with the default parameters of GNINA by specifying the extracted ligands to define binding sites. For docking of each compound into its target, the tool generates multiple poses with two scores: the one used to rank the poses of the ligand (i.e. CNNscore) and the one estimating the affinity of the docked complex (i.e. CNNaffinity). Since we are interested in the binding affinity of compounds against targets, we selected the pose with the highest affinity for each ligand-protein complex.

Next, we first assessed the discriminative ability of this docking tool by comparing the predicted affinity scores of compounds active towards a target with the scores of randomly sampled compounds for each structure. We chose Receiving Operating Characteristic (ROC) curve and Area Under the ROC Curve (AUC) as the evaluation metrics considering the active compounds as positive examples and randomly selected compounds as negative examples. These metrics measure whether this tool is able to distinguish between positive and negative classes. We also performed Mann-Whitney’s U test to check the discrimination between the compounds is significant or not. For 11 out of 12 proteins, the tool was able to discriminate between the active compounds and random ones. Since the active compounds and randomly selected ones cannot be distinguished for the structure with PDB code 2PJL, we excluded it from the rest of the experiments.

We next investigated whether the generated compounds are likely to
bind to the targets of interest. To this end, we compared these compounds with randomly selected compounds and the active compounds of the protein of interest. If two sets of compounds cannot be distinguished from one another, this suggests that these sets of compounds have a close binding affinity toward the target protein. On the other hand, one group is more or less likely to bind to the target than the other group depending on the AUC score if such groups can be discriminated. 

To determine the overall model performance, we computed the number of proteins for which the model can generate compounds that can be distinguished from randomly selected compounds and bind to the corresponding target as likely as or more likely than the known binders of that target. The results are reported in Table \ref{tbl:genvsrand}. The highest number of proteins (11), for which generated compounds are more likely to bind to a target of interest than randomly selected compounds were obtained with the compounds generated by the EncDecBase model with beam decoding. This result is consistent with the target level evaluation based on FCD scores and suggests that this model can generate chemical compounds with binding affinity to the target of interest.

\begin{table}[htbp]
\centering
\caption{Number of proteins in which generated compounds can be distinguished from random compounds for each model and decoding strategy.}
\label{tbl:genvsrand}
\begin{tabular}{cccc}
\hline
\textbf{Decoding}         %&  \textbf{Dataset}      
& \textbf{Model} & \multicolumn{1}{c}{\textbf{Generated vs Random}} & \textbf{Generated vs Active} \\  \hline
\multirow{3}{*}{Beam}     & %\multirow{3}{*}{100\%} &
EncDecBase     & 11                                               & 8                            \\
                          &                       % &
                          EncDecLM       & 8                                                & 6                            \\
                        %  &    
                        & T5             & 7                                                & 5                            \\ 
                        %   & \multirow{3}{*}{25\%}  & EncDecBase     & 7                                                & 6                            \\
                        %   &                        & EncDecLM       & 7                                                & 6                            \\
                        %   &                        & T5             & 4                                                & 3                            \\
                        \hline
\multirow{3}{*}{Sampling} %& \multirow{3}{*}{100\%} 
& EncDecBase     & 8                                                & 4                            \\
                          &                     %   &
                          EncDecLM       & 8                                                & 4                            \\
                         % &
                         & T5             & 2                                                & 1                            \\ 
                        %   & \multirow{3}{*}{25\%}  & EncDecBase     & 6                                                & 2                            \\
                        %   &                        & EncDecLM       & 6                                                & 2                            \\
                        %   &                        & T5             & 2                                                & 2    \\
                        \hline                       
\end{tabular}
\end{table}

EncDecBase model performed the best and generated compounds with binding affinity to a target of interest when decoded with beam search. However, the generated compounds must also be specific to the target to avoid undesired side effects caused by binding to others. To assess the target specificity, we selected 100 compounds for each protein among compounds generated for the other ten proteins and docked such compounds to the protein of interest. Then, we compared the binding affinity of such compounds to that of the following sets we compiled earlier: the compounds generated for the target of interest and the compounds active towards the target. The results are summarized in Table \ref{tbl:selectivity}. For reference, we also included the result of the generated compounds for the targets compared to the active compounds.

\begin{table}[htbp]
\centering
  \caption{Comparison of activity against target of interest between the active compounds, the generated ones for the target and the generated ones for the other proteins. Others refer to a random subset of the compounds generated for the other ten proteins.}
  \label{tbl:selectivity}

\begin{tabular}{c|cc|cc|cc}
\hline
\multirow{2}{*}{\textbf{Target}} & \multicolumn{2}{c|}{\textbf{Active vs Generated}} & \multicolumn{2}{c|}{\textbf{Active vs Others}} & \multicolumn{2}{c}{\textbf{Generated vs Others}} \\
                                 & \textbf{p value}         & \textbf{AUC}         & \textbf{p value}        & \textbf{AUC}       & \textbf{p value}          & \textbf{AUC}         \\ \hline
\textbf{1ERE}                    & 1.74E-05        & 0.79        & 1.73E-06                & 0.69               & 0.418           & 0.52                 \\
\textbf{2WO6}                    & 0.013                    & 0.33                 & 0.007                   & 0.6                & 0.001                     & 0.74                 \\
\textbf{4B6L}                    & 0.014                    & 0.33                 & 1.98E-13                & 0.87               & 1.81E-08                  & 0.9                  \\
\textbf{4I23}                    & 0.027                    & 0.36                 & 4.46E-08                & 0.72               & 2.11E-06                  & 0.83                 \\
\textbf{5K0K}                    & 0.332                    & 0.47                 & 2.84E-09                & 0.74               & 9.28E-05                  & 0.77                 \\
\textbf{5TUY}                    &  4.90E-06        & 0.83       & 2.77E-14                & 0.85               & 0.001                     & 0.73                 \\
\textbf{5V1B}                    & 2.28E-06                 & 0.17                 & 0.035                   & 0.58               & 4.76E-07                  & 0.85                 \\
\textbf{5XY1}                    & 0.018           & 0.65       & 1.56E-11                & 0.79               & 0.001                     & 0.72                 \\
\textbf{6LVL}                    & 0.389                    & 0.48                 & 4.50E-13                & 0.8                & 3.95E-06                  & 0.82                 \\
\textbf{6WJ5}                    & 4.58E-05                 & 0.22                 & 0.012                   & 0.59               & 2.69E-06                  & 0.82                 \\
\textbf{6Z1Q}                    & 0.454                    & 0.49                 & 0.053                   & 0.57               & \textit{0.175}            & 0.57   \\ \hline              
\end{tabular}
\end{table}

We observed that high AUC values were obtained when comparing the active compounds versus the compounds generated for others. This indicates that the docking tool is more likely to classify the compounds targeting others as nonbinders for the target of interest. By contrast, the generated compounds for only three targets (1ERE, 5TUY, 5XY1) were more likely to be classified as nonbinders. Additionally, the comparison between the compounds generated for the target of interest and those generated for others shows that these compound sets are significantly different for all targets except for 1ERE and 6Z1Q. Taken together, these results suggest that this model can generate target specific compounds for the majority of the targets.

\section{Discussion}
Our findings suggest that the proposed warm starting strategies to initialize models outperform the T5 model trained from scratch across decoding methods. The proposed method might be beneficial for other tasks in cheminformatics where labelled data are limited. Even in the presence of a large amount of data, it can be used to reduce the computational burden. However, we should also note that pretrained models used to initialize models require much more space than the baseline model. The results demonstrate that the model warm started with the one-stage strategy can generate target specific compounds for most of the set of novel proteins. Thus, it might be practical at the beginning of the drug design with little information regarding the target of interest.

Despite the potential benefits, the proposed model has certain limitations. The compounds generated by beam search are predicted to be able to bind to the protein of interest but have low diversity. The deterministic nature of the model combined with the variability of protein interactions hinders further generalizability of the model. To overcome this issue, one may incorporate stochastic latent variables into the model to improve the capability to learn the variability of interactions \citep{Lin2020,Arroyo}. Another direction for improving the diversity of compounds is to use ancestral sampling methods, which have been proposed recently and have been shown to generate high diversity samples in neural machine translation \citep{kool2020ancestral, Eikema}.

\section{Conclusion}
In this study, we proposed warm-start strategies to initialize models for target specific molecule generation. We compared two strategies: one stage strategy where the model is initialized with pretrained checkpoints and trained on targeted molecule generation (EncDecBase) and two-stage strategy where the model is first finetuned on molecular generation and then trained on targeted drug design (EncDecLM). To generate molecules, we employed beam search and sampling. We evaluated the generated molecules with metrics assessing the quality, diversity and performing docking. The results showed the efficacy of the warm-starting approach. The warm-started models obtained significantly better scores than the baseline model (i.e., T5) across decoding methods and performed on par with each other with respect to the benchmarking metrics. However, based on docking results, the one-stage warm-start strategy generated molecules likely to bind to the target of interest for more proteins, indicating that the model generalizes better than the two-stage strategy and the baseline. For future work, we plan to investigate adding stochastic latent variables into our models to increase the diversity of compounds similar to Variational Transformer models \citep{Lin2020, Arroyo}.

\section*{Acknowledgements}
We thank Asu Büşra Temizer and Taha Koulani for their help with evaluation and Riza Özçelik for his help in data preprocessing. TUBA-GEBIP Award of the Turkish Science Academy (to A.O.) is gratefully acknowledged. This work was supported by the Scientific and Technological Research Council of Turkey [Grant Numbers 119E133 and 2211-A to G.U.].

\textbf{Conflict of Interest:} E.O. is an employee of Roche AG.

\bibliographystyle{unsrtnat}
\bibliography{main}  %%% Uncomment this line and comment out the ``thebibliography'' section below to use the external .bib file (using bibtex) .

\begin{thebibliography}{48}
\providecommand{\natexlab}[1]{#1}
\providecommand{\url}[1]{\texttt{#1}}
\expandafter\ifx\csname urlstyle\endcsname\relax
  \providecommand{\doi}[1]{doi: #1}\else
  \providecommand{\doi}{doi: \begingroup \urlstyle{rm}\Url}\fi

\bibitem[Polishchuk et~al.(2013)Polishchuk, Madzhidov, and
  Varnek]{polishchuk2013estimation}
Pavel~G Polishchuk, Timur~I Madzhidov, and Alexandre Varnek.
\newblock Estimation of the size of drug-like chemical space based on gdb-17
  data.
\newblock \emph{Journal of Computer-Aided Molecular Design}, 27\penalty0
  (8):\penalty0 675--679, 2013.

\bibitem[Pe{\'o}n et~al.(2017)Pe{\'o}n, Naulaerts, and
  Ballester]{peon2017predicting}
Antonio Pe{\'o}n, Stefan Naulaerts, and Pedro~J Ballester.
\newblock Predicting the reliability of drug-target interaction predictions
  with maximum coverage of target space.
\newblock \emph{Scientific Reports}, 7\penalty0 (1):\penalty0 1--11, 2017.

\bibitem[Kramer et~al.(2007)Kramer, Sagartz, and Morris]{kramer2007application}
Jeffrey~A Kramer, John~E Sagartz, and Dale~L Morris.
\newblock The application of discovery toxicology and pathology towards the
  design of safer pharmaceutical lead candidates.
\newblock \emph{Nature Reviews Drug Discovery}, 6\penalty0 (8):\penalty0
  636--649, 2007.

\bibitem[Aumentado-Armstrong(2018)]{aumentado2018latent}
Tristan Aumentado-Armstrong.
\newblock Latent molecular optimization for targeted therapeutic design.
\newblock \emph{arXiv preprint arXiv:1809.02032}, 2018.

\bibitem[Skalic et~al.(2019)Skalic, Sabbadin, Sattarov, Sciabola, and {De
  Fabritiis}]{Skalic2019}
Miha Skalic, Davide Sabbadin, Boris Sattarov, Simone Sciabola, and Gianni {De
  Fabritiis}.
\newblock {From target to drug: Generative modeling for the multimodal
  structure-based ligand design}.
\newblock \emph{Molecular Pharmaceutics}, 16\penalty0 (10):\penalty0
  4282--4291, 2019.
\newblock ISSN 15438392.
\newblock \doi{10.1021/acs.molpharmaceut.9b00634}.

\bibitem[Moret et~al.(2020)Moret, Friedrich, Grisoni, Merk, and
  Schneider]{Moret2020}
Michael Moret, Lukas Friedrich, Francesca Grisoni, Daniel Merk, and Gisbert
  Schneider.
\newblock {Generative molecular design in low data regimes}.
\newblock \emph{Nature Machine Intelligence}, 2\penalty0 (3):\penalty0
  171--180, 2020.
\newblock ISSN 2522-5839.

\bibitem[Born et~al.(2021)Born, Manica, Cadow, Markert, Mill, Filipavicius,
  Janakarajan, Cardinale, Laino, and Mart{\'{i}}nez]{Born2021}
Jannis Born, Matteo Manica, Joris Cadow, Greta Markert, Nil~Adell Mill,
  Modestas Filipavicius, Nikita Janakarajan, Antonio Cardinale, Teodoro Laino,
  and Mar{\'{i}}a~Rodr{\'{i}}guez Mart{\'{i}}nez.
\newblock {Data-driven molecular design for discovery and synthesis of novel
  ligands: A case study on SARS-CoV-2}.
\newblock \emph{Machine Learning: Science and Technology}, 2\penalty0
  (2):\penalty0 25024, 2021.
\newblock ISSN 26322153.

\bibitem[Chenthamarakshan et~al.(2020)Chenthamarakshan, Das, Padhi, Strobelt,
  Lim, Hoover, Hoffman, and Mojsilovic]{Chenthamarakshan2020}
Vijil Chenthamarakshan, Payel Das, Inkit Padhi, Hendrik Strobelt, Kar~Wai Lim,
  Benjamin Hoover, Samuel~C Hoffman, and Aleksandra Mojsilovic.
\newblock Target-specific and selective drug design for covid-19 using deep
  generative models.
\newblock \emph{arXiv preprint arXiv:2004.01215}, 2020.

\bibitem[Renz et~al.(2020)Renz, Van~Rompaey, Wegner, Hochreiter, and
  Klambauer]{renz2020failure}
Philipp Renz, Dries Van~Rompaey, J{\"o}rg~Kurt Wegner, Sepp Hochreiter, and
  G{\"u}nter Klambauer.
\newblock On failure modes in molecule generation and optimization.
\newblock \emph{Drug Discovery Today: Technologies}, 2020.

\bibitem[Grechishnikova(2021)]{Grechishnikova2021}
Daria Grechishnikova.
\newblock {Transformer neural network for protein-specific de novo drug
  generation as a machine translation problem}.
\newblock \emph{Scientific Reports}, 11\penalty0 (1):\penalty0 1--13, 2021.
\newblock ISSN 20452322.

\bibitem[Devlin et~al.(2019)Devlin, Chang, Lee, and Toutanova]{Devlin2019}
Jacob Devlin, Ming~Wei Chang, Kenton Lee, and Kristina Toutanova.
\newblock {BERT: Pre-training of deep bidirectional transformers for language
  understanding}.
\newblock In \emph{NAACL HLT 2019 - 2019 Conference of the North American
  Chapter of the Association for Computational Linguistics: Human Language
  Technologies - Proceedings of the Conference}, volume~1, pages 4171--4186.
  Association for Computational Linguistics (ACL), 2019.
\newblock ISBN 9781950737130.

\bibitem[{\"O}z{\c{c}}elik et~al.(2021){\"O}z{\c{c}}elik, {\"O}zt{\"u}rk,
  {\"O}zg{\"u}r, and Ozkirimli]{ozccelik2021chemboost}
R{\i}za {\"O}z{\c{c}}elik, Hakime {\"O}zt{\"u}rk, Arzucan {\"O}zg{\"u}r, and
  Elif Ozkirimli.
\newblock Chemboost: A chemical language based approach for protein--ligand
  binding affinity prediction.
\newblock \emph{Molecular Informatics}, 40\penalty0 (5):\penalty0 2000212,
  2021.

\bibitem[Chithrananda et~al.(2020)Chithrananda, Grand, and
  Ramsundar]{Chithrananda}
Seyone Chithrananda, Gabriel Grand, and Bharath Ramsundar.
\newblock Chemberta: Large-scale self-supervised pretraining for molecular
  property prediction.
\newblock \emph{arXiv preprint arXiv:2010.09885}, 2020.

\bibitem[Filipavicius et~al.(2020)Filipavicius, Manica, Cadow, and
  Martinez]{Filipavicius2020}
Modestas Filipavicius, Matteo Manica, Joris Cadow, and Maria~Rodriguez
  Martinez.
\newblock {Pre-training protein language models with label-agnostic binding
  pairs enhances performance in downstream tasks}.
\newblock \emph{arXiv preprint arXiv:2012.03084}, 2020.

\bibitem[Rives et~al.(2021)Rives, Meier, Sercu, Goyal, Lin, Liu, Guo, Ott,
  Zitnick, Ma, et~al.]{rives2021biological}
Alexander Rives, Joshua Meier, Tom Sercu, Siddharth Goyal, Zeming Lin, Jason
  Liu, Demi Guo, Myle Ott, C~Lawrence Zitnick, Jerry Ma, et~al.
\newblock Biological structure and function emerge from scaling unsupervised
  learning to 250 million protein sequences.
\newblock \emph{Proceedings of the National Academy of Sciences}, 118\penalty0
  (15), 2021.

\bibitem[Elnaggar et~al.(2020)Elnaggar, Heinzinger, Dallago, Rihawi, Wang,
  Jones, Gibbs, Feher, Angerer, Steinegger, et~al.]{elnaggar2020prottrans}
Ahmed Elnaggar, Michael Heinzinger, Christian Dallago, Ghalia Rihawi, Yu~Wang,
  Llion Jones, Tom Gibbs, Tamas Feher, Christoph Angerer, Martin Steinegger,
  et~al.
\newblock Prottrans: towards cracking the language of life's code through
  self-supervised deep learning and high performance computing.
\newblock \emph{arXiv preprint arXiv:2007.06225}, 2020.

\bibitem[Weininger(1988)]{weininger1988smiles}
David Weininger.
\newblock Smiles, a chemical language and information system. 1. introduction
  to methodology and encoding rules.
\newblock \emph{Journal of chemical information and computer sciences},
  28\penalty0 (1):\penalty0 31--36, 1988.

\bibitem[Rothe et~al.(2020)Rothe, Narayan, and Severyn]{rothe2020leveraging}
Sascha Rothe, Shashi Narayan, and Aliaksei Severyn.
\newblock Leveraging pre-trained checkpoints for sequence generation tasks.
\newblock \emph{Transactions of the Association for Computational Linguistics},
  8:\penalty0 264--280, 2020.

\bibitem[Gilson et~al.(2016)Gilson, Liu, Baitaluk, Nicola, Hwang, and
  Chong]{Gilson2016}
Michael~K. Gilson, Tiqing Liu, Michael Baitaluk, George Nicola, Linda Hwang,
  and Jenny Chong.
\newblock {BindingDB in 2015: A public database for medicinal chemistry,
  computational chemistry and systems pharmacology}.
\newblock \emph{Nucleic Acids Research}, 44\penalty0 (D1):\penalty0
  D1045--D1053, 2016.
\newblock ISSN 13624962.

\bibitem[Polykovskiy et~al.(2020)Polykovskiy, Zhebrak, Sanchez-Lengeling,
  Golovanov, Tatanov, Belyaev, Kurbanov, Artamonov, Aladinskiy, Veselov,
  Kadurin, Johansson, Chen, Nikolenko, Aspuru-Guzik, and
  Zhavoronkov]{Polykovskiy2020}
Daniil Polykovskiy, Alexander Zhebrak, Benjamin Sanchez-Lengeling, Sergey
  Golovanov, Oktai Tatanov, Stanislav Belyaev, Rauf Kurbanov, Aleksey
  Artamonov, Vladimir Aladinskiy, Mark Veselov, Artur Kadurin, Simon Johansson,
  Hongming Chen, Sergey Nikolenko, Al{\'{a}}n Aspuru-Guzik, and Alex
  Zhavoronkov.
\newblock {Molecular Sets (MOSES): A benchmarking platform for molecular
  generation models}.
\newblock \emph{Frontiers in Pharmacology}, 11:\penalty0 1--19, 2020.
\newblock ISSN 16639812.

\bibitem[Raffel et~al.(2020)Raffel, Shazeer, Roberts, Lee, Narang, Matena,
  Zhou, Li, and Liu]{Raffel2019}
Colin Raffel, Noam Shazeer, Adam Roberts, Katherine Lee, Sharan Narang, Michael
  Matena, Yanqi Zhou, Wei Li, and Peter~J. Liu.
\newblock Exploring the limits of transfer learning with a unified text-to-text
  transformer.
\newblock \emph{Journal of Machine Learning Research}, 21\penalty0
  (140):\penalty0 1--67, 2020.
\newblock URL \url{http://jmlr.org/papers/v21/20-074.html}.

\bibitem[Mistry et~al.(2021)Mistry, Chuguransky, Williams, Qureshi, Salazar,
  Sonnhammer, Tosatto, Paladin, Raj, Richardson, et~al.]{mistry2021pfam}
Jaina Mistry, Sara Chuguransky, Lowri Williams, Matloob Qureshi, Gustavo~A
  Salazar, Erik~LL Sonnhammer, Silvio~CE Tosatto, Lisanna Paladin, Shriya Raj,
  Lorna~J Richardson, et~al.
\newblock Pfam: The protein families database in 2021.
\newblock \emph{Nucleic Acids Research}, 49\penalty0 (D1):\penalty0 D412--D419,
  2021.

\bibitem[Jansson-L{\"o}fmark et~al.(2020)Jansson-L{\"o}fmark, Hjorth, and
  Gabrielsson]{jansson2020does}
Rasmus Jansson-L{\"o}fmark, Stephan Hjorth, and Johan Gabrielsson.
\newblock Does in vitro potency predict clinically efficacious concentrations?
\newblock \emph{Clinical Pharmacology \& Therapeutics}, 108\penalty0
  (2):\penalty0 298--305, 2020.

\bibitem[Gao et~al.(2018)Gao, Fokoue, Luo, Iyengar, Dey, and
  Zhang]{gao2018interpretable}
Kyle~Yingkai Gao, Achille Fokoue, Heng Luo, Arun Iyengar, Sanjoy Dey, and Ping
  Zhang.
\newblock Interpretable drug target prediction using deep neural
  representation.
\newblock In \emph{IJCAI}, pages 3371--3377, 2018.

\bibitem[Cock et~al.(2009)Cock, Antao, Chang, Chapman, Cox, Dalke,
  et~al.]{10.1093/bioinformatics/btp163}
Peter J.~A. Cock, Tiago Antao, Jeffrey~T. Chang, Brad~A. Chapman, Cymon~J. Cox,
  Andrew Dalke, et~al.
\newblock {Biopython: freely available Python tools for computational molecular
  biology and bioinformatics}.
\newblock \emph{Bioinformatics}, 25\penalty0 (11):\penalty0 1422--1423, 03
  2009.
\newblock ISSN 1367-4803.
\newblock \doi{10.1093/bioinformatics/btp163}.
\newblock URL \url{https://doi.org/10.1093/bioinformatics/btp163}.

\bibitem[Rice et~al.(2000)Rice, Longden, and Bleasby]{rice2000emboss}
Peter Rice, Ian Longden, and Alan Bleasby.
\newblock Emboss: the european molecular biology open software suite.
\newblock \emph{Trends in genetics}, 16\penalty0 (6):\penalty0 276--277, 2000.

\bibitem[Sterling and Irwin(2015)]{sterling2015zinc}
Teague Sterling and John~J Irwin.
\newblock Zinc 15--ligand discovery for everyone.
\newblock \emph{Journal of chemical information and modeling}, 55\penalty0
  (11):\penalty0 2324--2337, 2015.

\bibitem[Woźniak et~al.(2018)Woźniak, Wołos, Modrzyk, Górski, Winkowski,
  Bajczyk, Szymkuć, Grzybowski, and Eder]{wozniak_linguistic_2018}
Michał Woźniak, Agnieszka Wołos, Urszula Modrzyk, Rafał~L. Górski, Jan
  Winkowski, Michał Bajczyk, Sara Szymkuć, Bartosz~A. Grzybowski, and Maciej
  Eder.
\newblock Linguistic measures of chemical diversity and the “keywords” of
  molecular collections.
\newblock \emph{Scientific Reports}, 8\penalty0 (1):\penalty0 7598, 2018.
\newblock ISSN 2045-2322.

\bibitem[Li and Fourches(2021)]{li2021smiles}
Xinhao Li and Denis Fourches.
\newblock Smiles pair encoding: A data-driven substructure tokenization
  algorithm for deep learning.
\newblock \emph{Journal of Chemical Information and Modeling}, 61\penalty0
  (4):\penalty0 1560--1569, 2021.

\bibitem[Asgari et~al.(2019)Asgari, McHardy, and
  Mofrad]{asgari_probabilistic_2019}
Ehsaneddin Asgari, Alice~C. McHardy, and Mohammad R.~K. Mofrad.
\newblock Probabilistic variable-length segmentation of protein sequences for
  discriminative motif discovery ({DiMotif}) and sequence embedding
  ({ProtVecX}).
\newblock \emph{Scientific Reports}, 9\penalty0 (1):\penalty0 3577, March 2019.
\newblock ISSN 2045-2322.

\bibitem[Vaswani et~al.(2017)Vaswani, Shazeer, Parmar, Uszkoreit, Jones, Gomez,
  Kaiser, and Polosukhin]{Vaswani2017}
Ashish Vaswani, Noam Shazeer, Niki Parmar, Jakob Uszkoreit, Llion Jones,
  Aidan~N. Gomez, {\L}ukasz Kaiser, and Illia Polosukhin.
\newblock {Attention is all you need}.
\newblock In \emph{Advances in Neural Information Processing Systems}, volume
  2017-Decem, pages 5999--6009, 2017.

\bibitem[Mahmood et~al.(2021)Mahmood, Mansimov, Bonneau, and Cho]{Mahmood2020}
Omar Mahmood, Elman Mansimov, Richard Bonneau, and Kyunghyun Cho.
\newblock Masked graph modeling for molecule generation.
\newblock \emph{Nature Communications}, 12\penalty0 (1):\penalty0 1--12, 2021.

\bibitem[Fabian et~al.(2020)Fabian, Edlich, Gaspar, Segler, Meyers, Fiscato,
  and Ahmed]{Fabian}
Benedek Fabian, Thomas Edlich, H{\'e}l{\'e}na Gaspar, Marwin Segler, Joshua
  Meyers, Marco Fiscato, and Mohamed Ahmed.
\newblock Molecular representation learning with language models and
  domain-relevant auxiliary tasks.
\newblock \emph{arXiv preprint arXiv:2011.13230}, 2020.

\bibitem[Szklarczyk et~al.(2019)Szklarczyk, Gable, Lyon, Junge, Wyder,
  Huerta-Cepas, et~al.]{Szklarczyk2019}
Damian Szklarczyk, Annika~L. Gable, David Lyon, Alexander Junge, Stefan Wyder,
  Jaime Huerta-Cepas, et~al.
\newblock {STRING v11: Protein-protein association networks with increased
  coverage, supporting functional discovery in genome-wide experimental
  datasets}.
\newblock \emph{Nucleic Acids Research}, 47\penalty0 (D1):\penalty0 D607--D613,
  2019.
\newblock ISSN 13624962.

\bibitem[Kim et~al.(2019)Kim, Chen, Cheng, Gindulyte, He, He, Li, Shoemaker,
  Thiessen, Yu, et~al.]{Kim2019}
Sunghwan Kim, Jie Chen, Tiejun Cheng, Asta Gindulyte, Jia He, Siqian He,
  Qingliang Li, Benjamin~A Shoemaker, Paul~A Thiessen, Bo~Yu, et~al.
\newblock Pubchem 2019 update: improved access to chemical data.
\newblock \emph{Nucleic acids research}, 47\penalty0 (D1):\penalty0
  D1102--D1109, 2019.

\bibitem[Liu et~al.(2019)Liu, Ott, Goyal, Du, Joshi, Chen, Levy, Lewis,
  Zettlemoyer, and Stoyanov]{liu2019roberta}
Yinhan Liu, Myle Ott, Naman Goyal, Jingfei Du, Mandar Joshi, Danqi Chen, Omer
  Levy, Mike Lewis, Luke Zettlemoyer, and Veselin Stoyanov.
\newblock Roberta: A robustly optimized bert pretraining approach.
\newblock \emph{arXiv preprint arXiv:1907.11692}, 2019.

\bibitem[Wolf et~al.(2020)Wolf, Debut, Sanh, Chaumond, Delangue, Moi, Cistac,
  et~al.]{Wolf2020}
Thomas Wolf, Lysandre Debut, Victor Sanh, Julien Chaumond, Clement Delangue,
  Anthony Moi, Pierric Cistac, et~al.
\newblock {Transformers: State-of-the-Art natural language processing}.
\newblock pages 38--45. Association for Computational Linguistics (ACL), 2020.

\bibitem[Landrum et~al.(2022)]{Landrum2016RDKit2016_09_4}
Greg Landrum et~al.
\newblock Rdkit: Open-source cheminformatics.
\newblock 2022.
\newblock URL \url{https://www.rdkit.org/}.

\bibitem[Preuer et~al.(2018)Preuer, Renz, Unterthiner, Hochreiter, and
  Klambauer]{Preuer2018}
Kristina Preuer, Philipp Renz, Thomas Unterthiner, Sepp Hochreiter, and
  G{\"{u}}nter Klambauer.
\newblock {Fr{\'{e}}chet ChemNet Distance: A metric for generative models for
  molecules in drug discovery}.
\newblock \emph{Journal of Chemical Information and Modeling}, 58\penalty0
  (9):\penalty0 1736--1741, 2018.
\newblock ISSN 15205142.

\bibitem[Bemis and Murcko(1996)]{Bemis1996}
Guy~W. Bemis and Mark~A. Murcko.
\newblock {The properties of known drugs. 1. Molecular frameworks}.
\newblock \emph{Journal of Medicinal Chemistry}, 39\penalty0 (15):\penalty0
  2887--2893, jul 1996.
\newblock ISSN 00222623.

\bibitem[Degen et~al.(2008)Degen, Wegscheid-Gerlach, Zaliani, and
  Rarey]{degen2008art}
J{\"o}rg Degen, Christof Wegscheid-Gerlach, Andrea Zaliani, and Matthias Rarey.
\newblock On the art of compiling and using'drug-like'chemical fragment spaces.
\newblock \emph{ChemMedChem: Chemistry Enabling Drug Discovery}, 3\penalty0
  (10):\penalty0 1503--1507, 2008.

\bibitem[Burley et~al.(2019)Burley, Berman, Bhikadiya, Bi, Chen, {Di Costanzo},
  Christie, Duarte, Dutta, Feng, et~al.]{Burley2019}
Stephen~K. Burley, Helen~M. Berman, Charmi Bhikadiya, Chunxiao Bi, Li~Chen,
  Luigi {Di Costanzo}, Cole Christie, Jose~M. Duarte, Shuchismita Dutta, Zukang
  Feng, et~al.
\newblock {Protein Data Bank: The single global archive for 3D macromolecular
  structure data}.
\newblock \emph{Nucleic Acids Research}, 47\penalty0 (D1):\penalty0 D520--D528,
  2019.
\newblock ISSN 13624962.
\newblock \doi{10.1093/nar/gky949}.

\bibitem[McNutt et~al.(2021)McNutt, Francoeur, Aggarwal, Masuda, Meli, Ragoza,
  Sunseri, and Koes]{McNutt2021}
Andrew~T. McNutt, Paul Francoeur, Rishal Aggarwal, Tomohide Masuda, Rocco Meli,
  Matthew Ragoza, Jocelyn Sunseri, and David~Ryan Koes.
\newblock {GNINA 1.0: molecular docking with deep learning}.
\newblock \emph{Journal of Cheminformatics}, 13\penalty0 (1):\penalty0 1--20,
  2021.
\newblock ISSN 17582946.

\bibitem[DeLano et~al.(2002)]{PyMOL}
Warren~L DeLano et~al.
\newblock Pymol: An open-source molecular graphics tool.
\newblock \emph{CCP4 Newsletter on Protein Crystallography}, 40\penalty0
  (1):\penalty0 82--92, 2002.

\bibitem[Lin et~al.(2020)Lin, Li, and Lin]{Lin2020}
Xiaoqian Lin, Xiu Li, and Xubo Lin.
\newblock {A review on applications of computational methods in drug screening
  and design}.
\newblock \emph{Molecules}, 25\penalty0 (6):\penalty0 1375, 2020.

\bibitem[Arroyo et~al.(2021)Arroyo, Postels, and Tombari]{Arroyo}
Diego~Martin Arroyo, Janis Postels, and Federico Tombari.
\newblock Variational transformer networks for layout generation.
\newblock In \emph{Proceedings of the IEEE/CVF Conference on Computer Vision
  and Pattern Recognition}, pages 13642--13652, 2021.

\bibitem[Kool et~al.(2020)Kool, van Hoof, and Welling]{kool2020ancestral}
Wouter Kool, Herke van Hoof, and Max Welling.
\newblock Ancestral gumbel-top-k sampling for sampling without replacement.
\newblock \emph{Journal of Machine Learning Research}, 21:\penalty0 47--1,
  2020.

\bibitem[Eikema and Aziz(2020)]{Eikema}
Bryan Eikema and Wilker Aziz.
\newblock Is map decoding all you need? the inadequacy of the mode in neural
  machine translation.
\newblock \emph{arXiv preprint arXiv:2005.10283}, 2020.

\end{thebibliography}

\appendix

\section{Appendix}

\counterwithin{figure}{section}
\setcounter{figure}{0}    

\subsection{Data Split Sequence Similarities}
\begin{figure}[h]
\centering
  \includegraphics[width=0.6\linewidth]{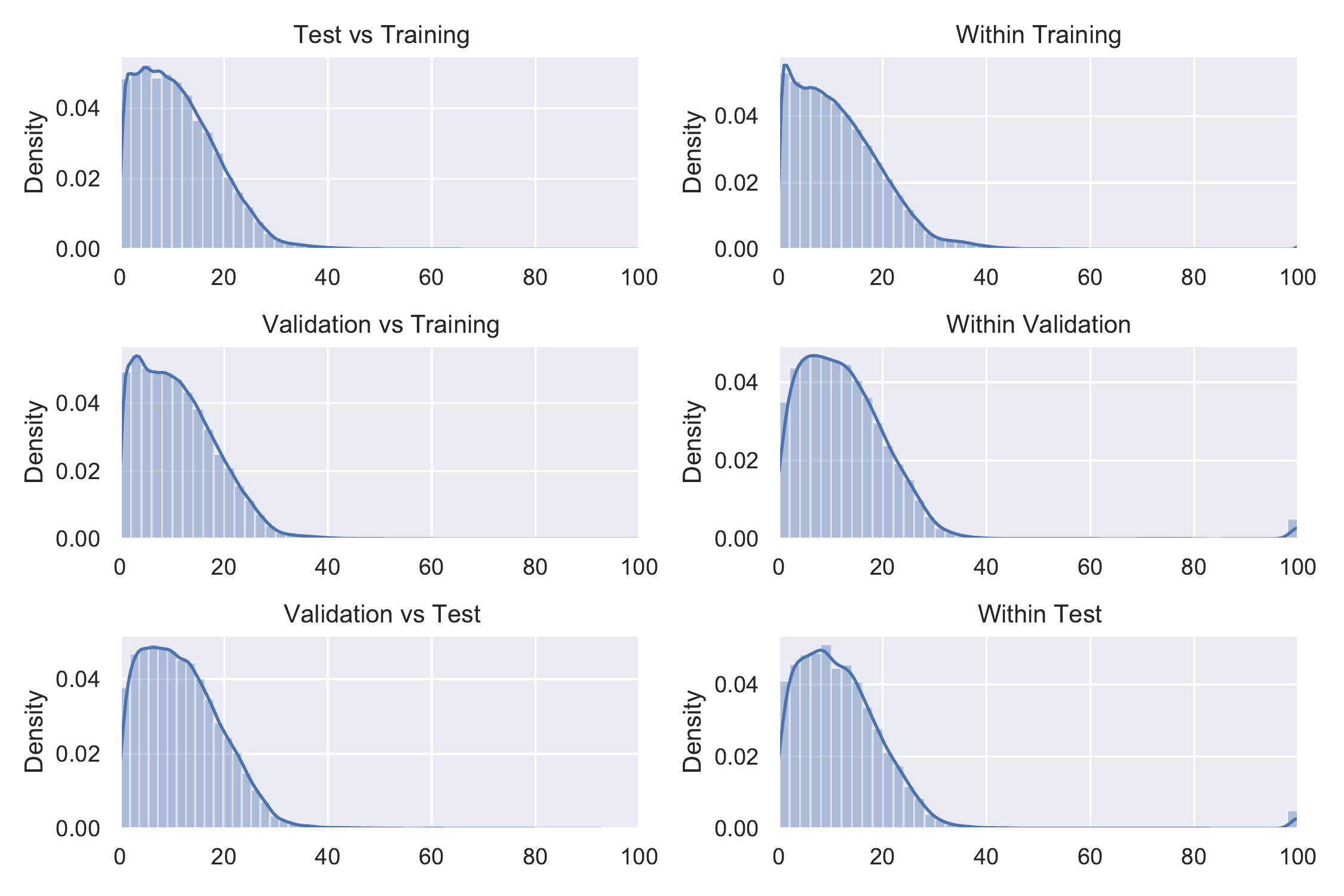}
  \caption{Distribution of protein sequence similarities between and within datasets (training, validation, test)} 
  \label{fgr:seq_sim}
\end{figure}

Distributions of sequence similarities between and within data splits are shown in Figure \ref{fgr:seq_sim}. 

\subsection{Protein Level Score Distribution}

\begin{figure}
     \centering
     \begin{subfigure}[b]{0.48\textwidth}
         \centering
         \includegraphics[width=\textwidth]{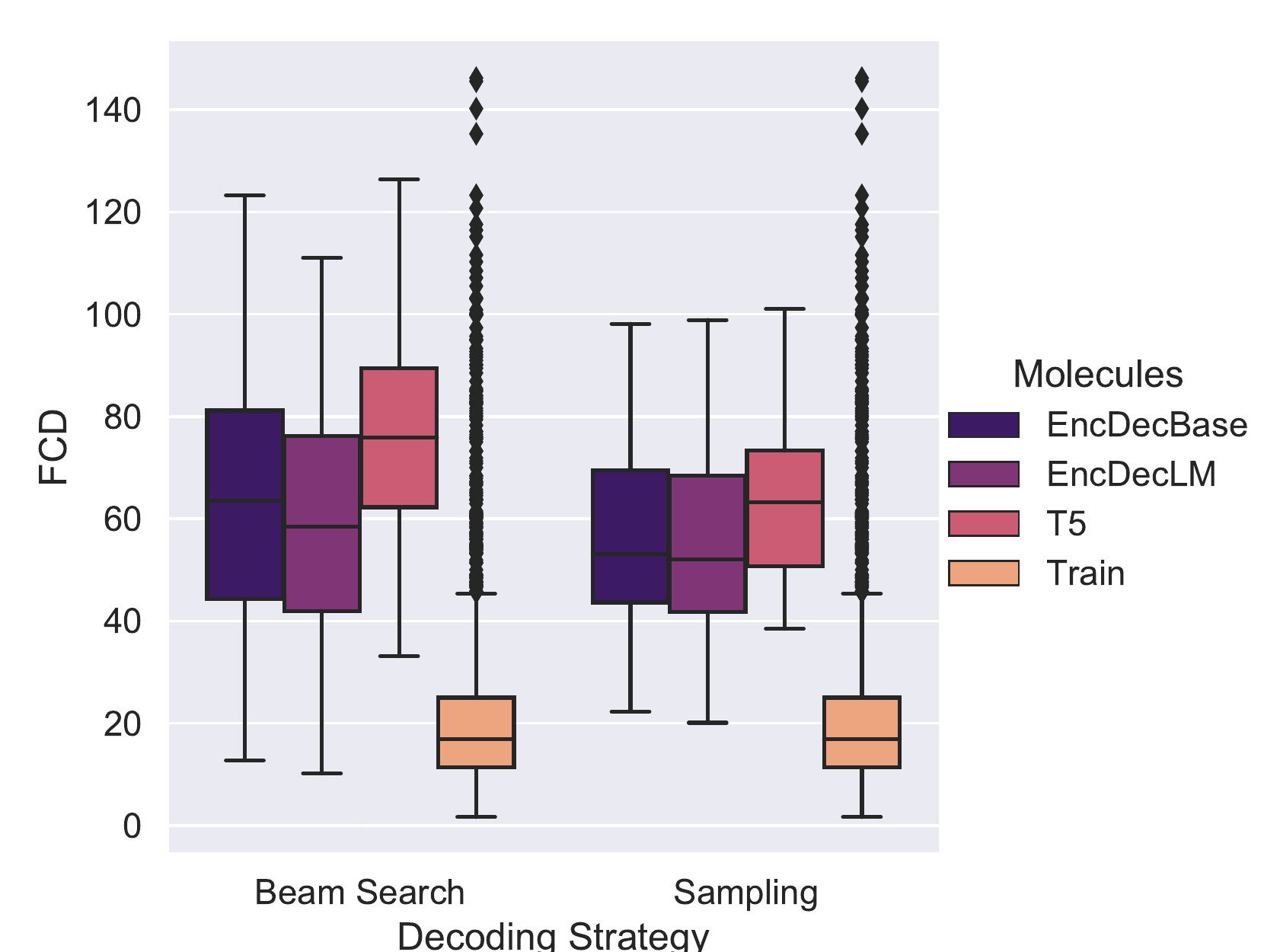}
         \caption{FCD distribution}
         \label{fgr:fcd2}
     \end{subfigure}
     \hfill
     \begin{subfigure}[b]{0.48\textwidth}
         \centering
         \includegraphics[width=\textwidth]{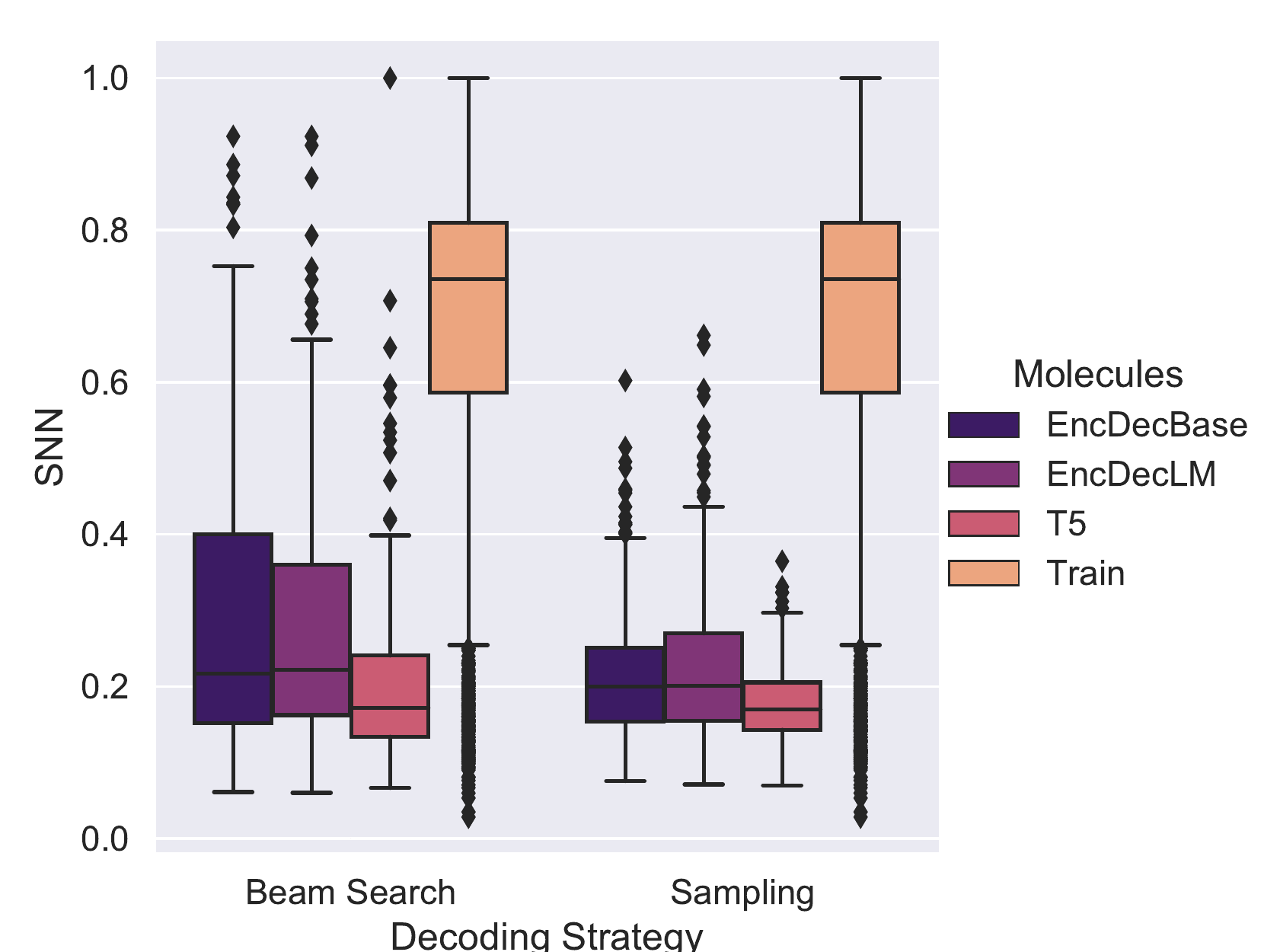}
         \caption{SNN distribution}
         \label{fgr:snn}
     \end{subfigure}
        \caption{Score distributions of chemical compounds generated for the test proteins across models and decoding strategies along with the distribution of the scores within interacting compounds of the train proteins.}
        \label{fig:three graphs}
\end{figure}

Generated molecules were also assessed at protein level. For each protein in the test set, we computed Fréchet ChemNet distance (FCD) \citep{Preuer2018} between the compounds generated for this protein and the compounds interacting with this protein (i.e. those labeled as active). For reference, we computed FCD scores for the proteins in training set by randomly selecting 20 interacting compounds and computing FCD between the selected chemical compounds and the remaining active compounds towards the target protein. Figure \ref{fgr:fcd2} shows the distribution of FCD scores across the decoding strategies. We also followed the same procedure to compute nearest neighbour Tanimoto similarity (SNN) scores for the proteins. The distributions of SNN scores are shown in Figure \ref{fgr:snn}. 
 
% \begin{figure}[h]
% \centering
%   \includegraphics[width=0.48\textwidth]{figures/target_FCD_box.pdf}
%     \caption{FCD score distribution of chemical compounds generated for the test proteins across models and decoding strategies along with the distribution of the scores within interacting compounds of the train proteins.} 
%   \label{fgr:fcd2}
% \end{figure}

% \begin{figure}[h]
% \centering
%   \includegraphics[width=0.48\textwidth]{figures/target_SNN_box.pdf}
%   \caption{SNN distribution of chemical compounds generated for the test proteins across models and decoding strategies along with the distribution of the scores within interacting compounds of the train proteins.} 
%   \label{fgr:snn}
% \end{figure}
\end{document}